\documentclass[runningheads]{llncs}

\usepackage{eccv}

\usepackage{eccvabbrv}
\usepackage[table]{xcolor}
\usepackage{graphicx}
\usepackage{booktabs}

\usepackage{multirow,makecell,hhline,amsfonts,arydshln,array,xcolor,xspace}
\usepackage{wrapfig}

\usepackage{algorithm}
\usepackage{algorithmic}

\usepackage[accsupp]{axessibility}  

\usepackage[pagebackref,breaklinks,colorlinks,citecolor=eccvblue]{hyperref}

\usepackage{orcidlink}

\begin{document}

\title{MVPruner: Dynamic Token Pruning for Accelerating Multi-view Vision-Language Models in Autonomous Driving} 

\titlerunning{MVPruner: Dynamic Token Pruning for Accelerating Multi-view VLMs}


\author{Nan Yang\inst{1}\orcidlink{0009-0007-0201-9388} \and
Zhanwen Liu\inst{1}\thanks{Corresponding author.}\orcidlink{0000-0002-8823-0833} \and
Linfeng Zhang\inst{2}\orcidlink{0000-0002-3341-183X} \and
Shangyu Xie\inst{1}\orcidlink{0009-0006-4232-7249} \and
Yang Wang\inst{1}\orcidlink{0000-0002-6259-6044} \and
Wenzhuo Zhou\inst{1} \and
Xiangmo Zhao\inst{1}}

\authorrunning{N.~Yang et al.}

\institute{School of Information Engineering, Chang'an University, China
\\
 \and
School of Artificial Intelligence, Shanghai Jiaotong University, China\\
\email{\{ynan,zwliu,ywang120\}@chd.edu.cn}
\email{zhanglinfeng@sjtu.edu.cn}}

\maketitle

\begin{abstract}
Vision-Language Models (VLMs) improve generalization and interpretability in autonomous driving but suffer from efficiency issues due to long visual token sequences, particularly in standard multi-view settings. Existing token pruning methods employ fixed pruning rate allocation and static importance metrics, ignoring dynamic inter-view importance differences and the evolving information importance during inference. Our analysis reveals that multi-view VLMs inherently encode task-related view priors in deeper layers and exhibit dynamic information requirements. Motivated by these findings, we propose MVPruner, a two-stage adaptive token pruning method that aligns pruning behavior with the model's dynamic information requirements. The first stage allocates pruning budgets based on the information diversity of each view, and retains tokens with consistent contribution across stages, ensuring semantic representational capacity. The second stage allocates budgets and selects tokens guided by instruction text to guarantee task alignment. Experimental results on four benchmarks demonstrate the superior performance of our method. For example, DriveMM equipped with MVPruner achieves 87.3\% reduction in FLOPs, 4.97× speedup in prefilling phase while retaining 98.5\% accuracy on DriveLM benchmark.

\end{abstract}

\section{Introduction}

\label{Introduction}

Vision-Language Models (VLMs) \cite{liu2023visual,bordes2024introduction,li2024llava} have shown promise for achieving interpretable and generalizable perception and decision-making in autonomous driving \cite{mao2023gpt,zhou2024vision,jiang2025survey,huang2025vlm,wang2025omnidrive,wei2026monocular}. However, to ensure comprehensive environmental perception, autonomous driving systems typically rely on multi-view camera inputs \cite{caesar2020nuscenes,liu2023multi,sun2020scalability,sima2024drivelm,ishaq2025drivelmm}. The high-resolution multi-view images inevitably generate excessive visual token sequences, leading to substantial computational overhead and inference latency \cite{chen2024image,zhang2024sparsevlm}, which severely hinders the deployment in latency-sensitive driving scenarios \cite{xiong2025prune2drive}.

\begin{figure}[tbp]
\centering
\includegraphics[scale=0.45]{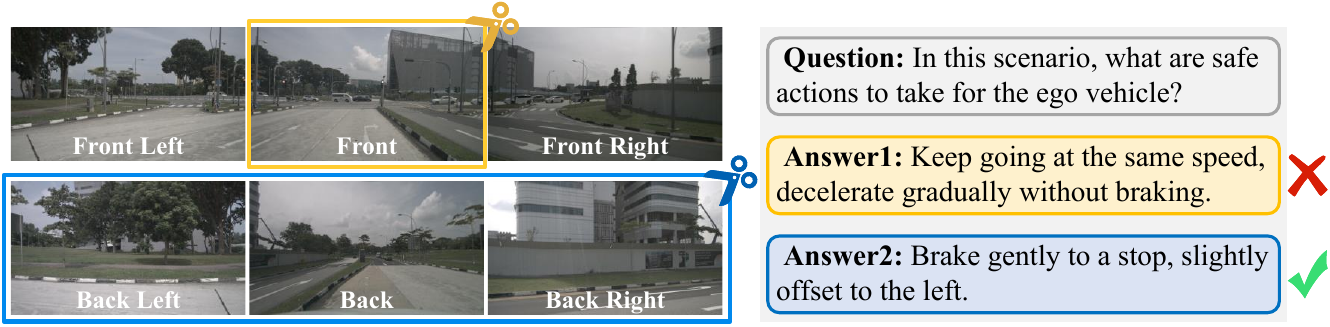}
\caption{Removing \textcolor[RGB]{0,111,191}{three unimportant views} does not affect the decision, while removing only the \textcolor{orange}{critical front view} leads to an \textcolor{red}{incorrect} decision, highlighting the inconsistent contributions across views.}
\label{fig1}
\end{figure}

To improve inference efficiency, visual token pruning has emerged as an effective strategy for eliminating redundant information \cite{liu2025video,wen2025token,yang2025smamba,shao2025tokens}. Although pruning techniques for VLMs have achieved preliminary progress, existing methods still face several critical challenges in multi-view autonomous driving scenarios: (1) Most existing methods are designed for single-view settings and apply uniform pruning ratios across views when handling multi-view inputs, ignoring the varying contributions of different views \cite{chen2024image,bolya2022token,zhang2024sparsevlm}. \cref{fig1} illustrates this effect: in braking scenarios triggered by a red light ahead, removing rear views does not affect the decision, whereas pruning the front view containing the key red light cue leads to incorrect and unsafe behavior. Although recent work (\eg, Prune2Drive \cite{xiong2025prune2drive}) allocates view-specific pruning budgets via offline hyperparameter search, such static strategies are sensitive to scene variations and cannot adapt to varying driving tasks and instructions. (2) Widely used token pruning methods rely on static importance evaluation metrics or single-stage pruning strategies, failing to account for the model’s dynamic information requirements and consequently resulting in suboptimal preservation of multi-view information.

\begin{figure}[tbp]
\centering
\includegraphics[scale=0.433]{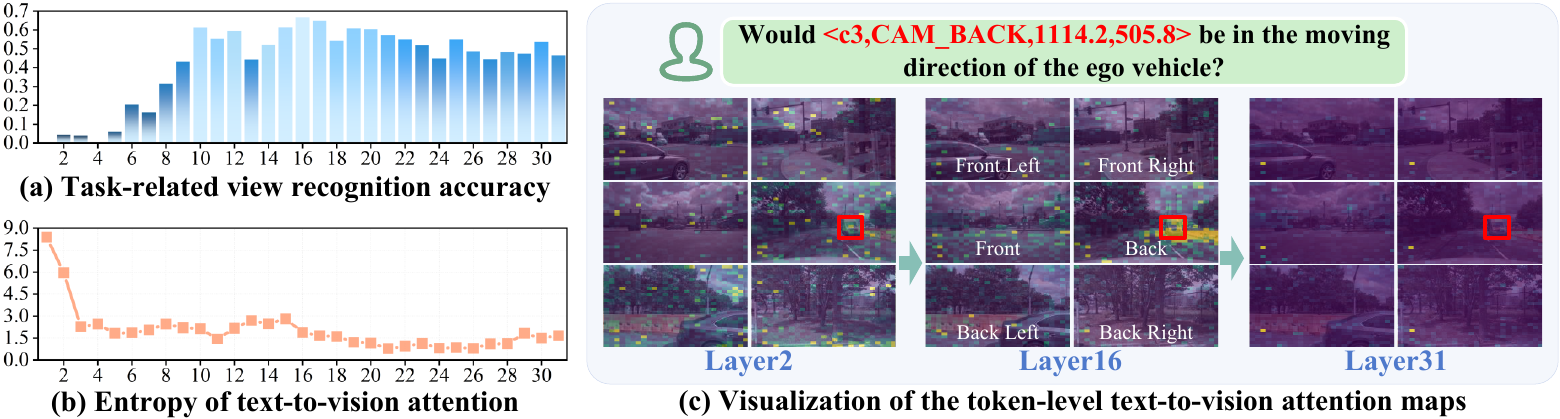}
\caption{\textbf{(a) Task-related view recognition accuracy across layers.} Higher accuracy indicates clearer identification of task-related views at this layer. \textbf{(b) Entropy of text-to-vision attention.} High entropy reflects diffuse global attention, while low entropy indicates concentrated local focus. \textbf{(c) Visualization of text-to-vision attention maps across layers.} Details are provided in Sec. \ref{main:analyze}.}
\label{fig2}
\end{figure}

To address these challenges, we first design a task-related view recognition experiment to quantitatively evaluate the ability of multi-view VLMs to recognize critical views across different layers. As shown in \cref{fig2}(a), the accuracy is low in shallow layers, increases substantially in intermediate layers where it reaches its peak. This result indicates that {\textbf{VLMs inherently encode task-related view priors during deeper inference}}. Furthermore, we investigate the underlying mechanisms behind this layer-wise discrepancy. As shown in Figs. \ref{fig2}(b) and (c), {\textbf{VLMs exhibit hierarchical attention patterns}} that shift from attending to broad visual context for global scene understanding to focusing on task-relevant local regions, and {\textbf{dynamic information requirements}} that transition from semantic diversity-driven processing to task relevance-driven processing.

The above findings provide principled guidance on \textit{{which stages to prune, how to allocate pruning budgets, and which tokens to retain}}. Based on these insights, we propose \textbf{MVPruner}, a two-stage adaptive token pruning framework for multi-view VLMs that aligns budget allocation and token selection with the model’s dynamic information requirements. In the first stage, inspired by the model’s broad attention to all visual information in shallow layers, we design the Diversity-aware Ratio Allocation strategy that measures intra-view feature diversity to assess view importance and allocate larger budgets to information-rich views. Meanwhile, a Cross-stage Contribution-aware Token Selection mechanism is proposed to retain tokens that consistently contribute across different layers, satisfying the dynamic information requirements during inference. In the second stage, the textual instructions can effectively guide view-importance estimation in deeper layers. Therefore, we introduce the Instruction-aware Ratio Allocation strategy that measures semantic relevance between visual and instruction tokens, assigning larger budgets to task-related views. Finally, we select tokens guided by instruction text to enhance task-alignment capabilities. Experiments on the image-based multi-view autonomous driving benchmarks: DriveLMM-o1\cite{ishaq2025drivelmm}, DriveLM \cite{sima2024drivelm}, and MAPLM \cite{cao2024maplm}, as well as video-based benchmark STSnu \cite{fruhwirth2025stsbench}  demonstrate that our method can consistently accelerate inference while effectively maintaining reasoning accuracy, exhibiting strong generalization in autonomous driving. Our contribution can be summarized as follows:

(1) Our analysis reveals that multi-view VLMs allocate more pronounced attention to task-related views in deeper layers and exhibit dynamic information requirements, providing insights to guide method design.

(2) We propose MVPruner, which dynamically adjusts pruning ratios across different views in an online manner and adaptively retains important tokens according to the model’s dynamic information requirements.

(3) Experimental results demonstrate that our method achieves a superior efficiency–performance trade-off. With 10\% of visual tokens retained, our method maintains accuracy of 98.0\%, 98.5\%, and 94.5\%, outperforming the second-best method by 2.6\%, 0.8\% and 3.5\%, achieving $3.80\times$, $4.97\times$ and $3.95\times$ speedup in the prefilling stage on DriveLMM-o1, DriveLM and MAPLM, respectively.

\section{Related Work}
\label{Related Work}
\subsection{VLMs for Autonomous Driving}
By jointly modeling visual inputs and language instructions within a shared embedding space, VLMs enhance reasoning, interpretability, and generalization in autonomous driving, supporting complex driving tasks \cite{kim2018textual,kim2024openvla,jiang2024senna,shao2024lmdrive,zhou2024vision,chen2024end}. Early efforts including GPT-Driver \cite{mao2023gpt}, demonstrate that frozen VLMs can process driving tasks while simultaneously producing human-readable rationales. Recently, DriveLM \cite{sima2024drivelm} and DriveLMM-o1 \cite{ishaq2025drivelmm} have respectively introduced a graph-based visual question answering dataset and a step-by-step reasoning dataset to enhance the driving-scene understanding of VLMs. DriveMM \cite{huang2024drivemm} designs a general VLM to process diverse data inputs and perform a broad spectrum of tasks. However, the long visual token sequences, especially for multi-view inputs, lead to significant inference latency and limit their practical applications.

\subsection{Visual Token Pruning for VLMs}
Recently, token pruning methods have attracted increasing attention as an effective acceleration strategy for VLMs \cite{rao2021dynamicvit,wen2025token}. These methods are generally categorized into two groups \cite{shao2025tokens}: (1) attention-based methods, which prune unimportant tokens via attention scores \cite{liang2022not,chen2024image,zhang2024sparsevlm,xing2024pyramiddrop}; (2) similarity-based methods \cite{bolya2022token,alvar2025divprune,wen2025stop}, which either merge similar tokens or maximize diversity of retained tokens. However, they mainly focus on single-view inputs and overlook the varying redundancy across views. Although recent work, Prune2Drive \cite{xiong2025prune2drive}, allocates pruning budgets across views through offline hyperparameter search, it incurs additional computational overhead and exhibits limited generalization capability. Moreover, these methods rely on static importance metrics or single-stage pruning, overlooking the model’s cross-layer dynamic information requirements.

\section{Method}
\label{Method}

\subsection{Multi-view Vision-Language Models}
A multi-view VLM \cite{sima2024drivelm,ishaq2025drivelmm} typically processes a pair of inputs, denoted as $(V, T)$, where $V$ is the multi-view image input and $T$ is the text input which is composed of system text and instruction text. The visual input $V\in\mathbb{R}^{P\times H \times W \times 3}$ is converted to visual tokens $X^{v}\in\mathbb{R}^{P\times M \times D}$ using a vision encoder and a projector layer, where $P$ is the number of views, $M$ is the number of tokens within each view and $D$ is the dimension of the feature. The text input is mapped to textual tokens $X^{t}\in\mathbb{R}^{N \times D}$ using a text encoder, where $N$ is the number of textual tokens. Then, they are fed to an LLM to generate the prediction.

\begin{figure}[tbp]
\centering
\includegraphics[scale=0.4]{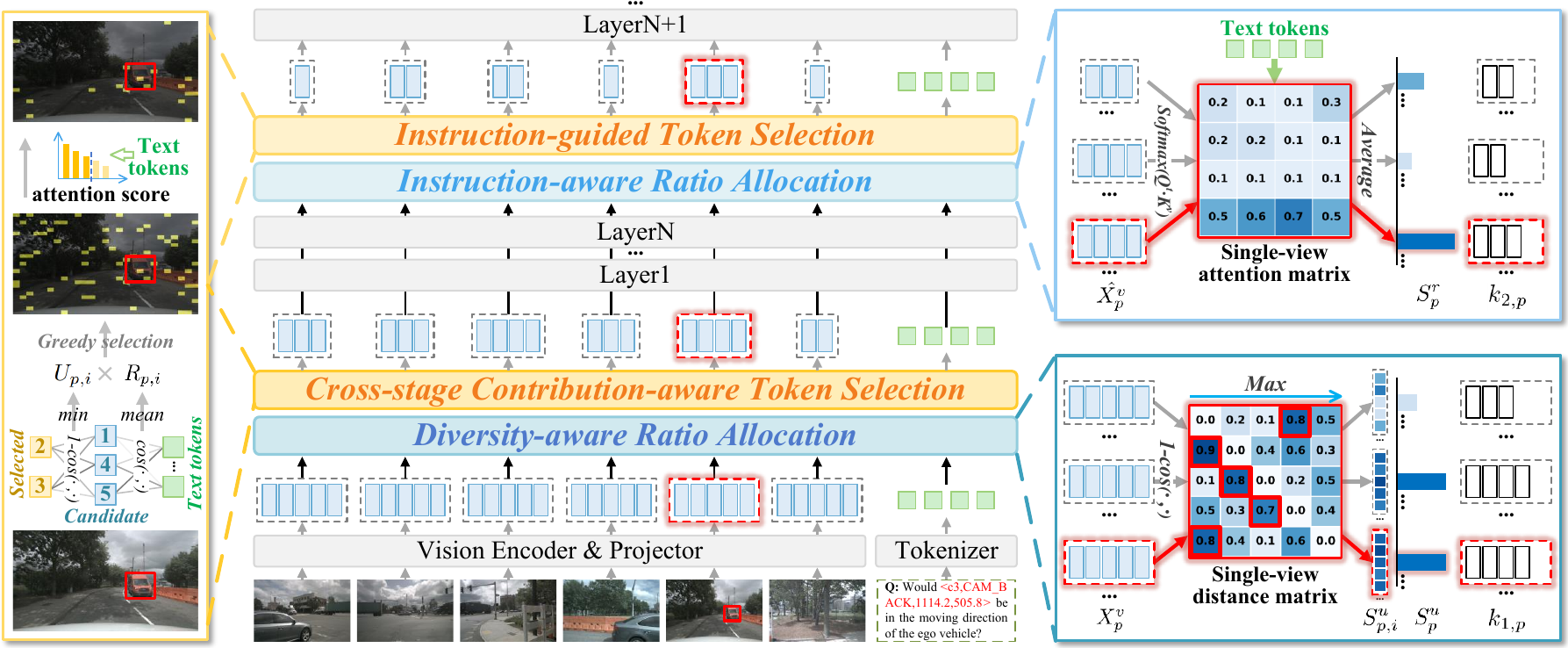}
\caption{\textbf{Overall framework of MVPruner.} In the shallow layer, MVPruner adjusts budget based on intra-view information diversity and selects tokens that consistently contribute across different layers. In the deeper layer, MVPruner allocates the budget via the semantic similarity between each view and the instruction text, and retains tokens with high cross-modal attention scores.}
\label{fig:framework}
\end{figure}

\subsection{Key Observations}
\label{main:analyze}

In autonomous driving tasks, textual instructions often exhibit explicit spatial specificity (\eg, “Would the object in the back-view camera be in the moving direction?”). These samples enable us to quantitatively analyze whether multi-view VLMs can inherently identify instruction-referenced views. To investigate this, we design a task-related view recognition experiment. Specifically, we evaluate two representative models, DriveMM \cite{huang2024drivemm} and DriveLMM-o1 \cite{ishaq2025drivelmm}, by extracting view-specific subsets from the DriveLM \cite{sima2024drivelm} and DriveLMM-o1 datasets. For each decoder layer, we compute the cross-modal attention scores between the instruction and visual tokens of each view. An identification is deemed correct if the view yielding the maximum score aligns with the instruction-referenced view. As shown in \cref{fig2}(a), as the layers deepen, the accuracy gradually increases and reaches its peak in the middle layers. This finding indicates that \textbf{cross-modal attention patterns in the deeper layers of VLMs effectively identify task-relevant views}.

Furthermore, to examine layer-wise differences in recognition accuracy, we analyze the cross-layer evolution of attention patterns between visual and textual modalities. Specifically, we compute the entropy of text-to-vision attention at each layer to measure the distribution of attended visual information. As shown in \cref{fig2}(b), entropy is initially high, decreases rapidly with fluctuations, and further declines in deep layers. This trend reflects a shift in the model’s attention patterns, consistent with the attention maps in \cref{fig2}(c). \textbf{In early layers}, textual semantics interact with broad, global visual context to establish initial scene understanding and coarse cross-modal alignment. \textbf{In intermediate layers}, the model progressively concentrates on task-relevant visual regions. \textbf{In deep layers}, visual information is further aggregated and exhibits consistently high attention at fixed spatial locations within each view. This finding suggests that \textbf{reasoning in VLMs exhibits dynamic information requirements that transition from semantic diversity-driven to task relevance-driven.}

\subsection{Framework Overview}
Inspired by the above findings, we propose MVPruner, a two-stage pruning framework that transitions from structural summarization to semantic focusing, aligning with the model’s dynamic information requirements, as shown in \cref{fig:framework}. Specifically, in the first stage, the Diversity-aware Ratio Allocation strategy is employed in the shallow layer to assign pruning rates based on differences in information diversity across views. Meanwhile, the Cross-stage Contribution-aware Token Selection mechanism retains tokens that consistently contribute across different stages. In the second stage, the Instruction-aware Ratio Allocation strategy is applied in the deeper layer, where pruning rates are determined by the task relevance of each view. Finally, redundant tokens are further removed guided by instruction.

\subsection{Diversity-aware Ratio Allocation}

In shallow layers, attention is relatively uniformly distributed across all views, reflecting the reasoning process driven by semantic diversity. This observation motivates us to start from the information richness within each view, preserving more information for views containing more diverse content, supplying the deeper layers with a more comprehensive visual context. To this end, we propose the Diversity-aware Ratio Allocation (DRA) strategy that dynamically assigns the pruning budget based on the measured feature diversity of each view, maximizing the diversity and completeness of visual representations during compression.
 
Firstly, we compute the semantic distances between all token pairs $\left(X_{p, i}^{v}, X_{p, j}^{v}\right)$ within each view $p$, $\left(i, j\right)\in M,i \neq j$ to estimate feature diversity. The distance $d_{p}(\cdot, \cdot)$ is defined as:
\begin{equation}
d_{p}\left(X_{p, i}^{v}, X_{p, j}^{v}\right) = 1 - \frac{X_{p, i}^{v} \cdot X_{p, j}^{v}}{\left\|X_{p, i}^{v}\right\| \left\|X_{p, j}^{v}\right\|}.
\label{eq:distance}
\end{equation}

For each token, the minimum pairwise distance is used as its uniqueness score $S_{p,i}^u$, where a higher score indicates that the token is more unique relative to others in the same view:
\begin{equation}
S_{p,i}^u = min\left(d_{p}(X_{p, i}^{v}, X_{p, 1:M}^{v})\right).
\label{eq:min}
\end{equation}

Subsequently, the diversity score of a view $S_{p}^u= \frac{1}{M} \sum_{i=1}^{M} S_{p,i}^u$ is obtained by averaging the uniqueness scores of all its tokens, with a higher score indicating a greater feature density compared to other views. The retention ratio $r_{1,p}$ and number $k_{1,p}$ for each view is determined based on the normalized $S_{p}^u$:
\begin{equation}
r_{1,p} = r_{1} \times P \times S_{p}^u,\quad 
k_{1,p}=r_{1,p}\times M,
\label{eq:allcate1}
\end{equation}
where $r_{1}$ is the average retention ratio for the first-stage pruning.

\subsection{Cross-stage Contribution-aware Token Selection}
Tokens retained in shallow layers should support task-agnostic global semantic modeling while providing a foundation for task-relevant representation extraction in deeper layers. To this end, we propose the Cross-stage Contribution-aware Token Selection (CCTS) mechanism that jointly evaluates a token's immediate contribution to the current stage and its potential contribution to subsequent stages, aligning selection with the model's dynamic information requirements.

Specifically, we formulate the token selection in shallow layers as an optimization problem aimed at maximizing cross-stage contribution. In this formulation, immediate contribution is quantified by semantic uniqueness within visual space, while future contribution is estimated through task relevance which is computed as the cosine similarity between visual and instruction tokens. An iterative greedy selection strategy is then employed to efficiently search for the optimal token subset.

Firstly, the token with the highest task relevance is selected to initialize the set of selected tokens. Subsequently, the minimum semantic distance between a candidate token and the selected set is computed using \cref{eq:min} and used as the semantic uniqueness metric, denoted by $U_{p,i}$. This metric is further weighted by the task relevance score $R_{p,i}$, enabling a comprehensive assessment of the token’s cross-stage contribution $Imp_{p,i}$:
\begin{equation}
Imp_{p,i}=U_{p,i} \times R_{p,i}.
\end{equation}

After that, the token with the highest cross-stage contribution is appended to the selected set. By iterating this process until the target token budget is reached, this mechanism maximizes cross-stage information utility. The algorithm flow is detailed in \cref{alg}.

\begin{algorithm}[!t]
\footnotesize
\caption{Cross-stage Contribution-aware Token Selection} 
\label{alg:token_pruning}
\begin{algorithmic}[1]
    \REQUIRE Visual tokens $X^{v}_{p}\in\mathbb{R}^{M \times D}$ of view $p$, candidate index set $\mathcal{C} \subseteq \{1,\dots,M\}$, target token number $\mathcal{K}=k_{1,p}$
    \ENSURE  Selected index set $\mathcal{S}$, with $|\mathcal{S}| = \mathcal{K}$
    \STATE \textbf{{Phase 1: Initialization}}
    \STATE Compute distance matrix: ${D}_{p,i,j} \gets d_{p}\left(X_{p, i}^{v}, X_{p, j}^{v}\right)$
    \STATE Compute semantic relevance: ${R}_{p,i} \gets cosine\left(X_{p, i}^{v}, X_{I}^{t}\right)$
    \STATE Select token with the highest semantic relevance: $i^\star \gets \arg\max ({R}_{p,i})$
    \STATE Initialize selected token list: $\mathcal{S} \gets \{i^\star\}$

    \STATE \textbf{{Phase 2: Iteratively add the subsequent tokens}}
    \WHILE{$|{\mathcal{S}}| < \mathcal{K}$}
    \FOR{each $j \in \mathcal{C} \setminus {\mathcal{S}}$}
        \STATE Uniqueness score: $U_{p,j} \gets \min_{i \in {\mathcal{S}}} D_{p,i,j}$
        \STATE Cross-stage contribution: $Imp_{p,j} \gets U_{p,j} \times R_{p,j}$
    \ENDFOR
    \STATE $j^\star \gets \arg\max_{j \in \mathcal{C} \setminus {\mathcal{S}}} (Imp_{p,j})$
    \STATE ${\mathcal{S}} \gets {\mathcal{S}} \cup \{j^\star\}$
    \ENDWHILE
    \RETURN$\mathcal{S}$
\end{algorithmic}
\label{alg}
\end{algorithm}

\subsection{Instruction-aware Ratio Allocation}
 
In deeper layers, attention is focused on local task-related regions, and the model can distinguish the task relevance of different views, reflecting the reasoning process driven by task relevance. Motivated by this, we design the Instruction-aware Ratio Allocation (IRA) strategy for the second stage, which adaptively allocates pruning budget based on semantic relevance between visual and instruction tokens, retaining more critical information for task-related views.

Specifically, we extract the attention matrix between the visual tokens and textual instruction tokens, and compute the mean attention score of all tokens within each view as its semantic relevance score $S_{p}^r$. A higher score indicates that the view exhibits greater task relevance than other views:
\begin{equation}
S_{p,i}^r =\frac{1}{\tilde{N}}\sum_{j=1}^{\tilde{N}}A_{i,j}^{p},\quad
S_{p}^r = \frac{1}{k_{1,p}}\sum_{i=1}^{k_{1,p}}S_{p,i}^r,
\label{eq:sematic}
\end{equation}
where $\tilde{N}$ is number of instruction tokens, $S_{p,i}^r$ is attention score of $i$-th token in view $p$, $A^{p}$ is attention matrix between view $p$ and instruction tokens.

After that, we reallocate the remaining token budget across views using a compensatory normalization scheme:
\begin{equation}
r_{2,p} = r_{2} \times \left(1 + S_{p}^r - S_{p}^u \right),\quad k_{2,p}=r_{2,p}\times k_{1,p},
\label{eq:allcate2}
\end{equation}
where $r_{2}$ denotes the average retention ratio for the second-stage pruning, and 
$S_{p}^r - S_{p}^u$ represents the relative deviation from the diversity score. $r_{2,p}$ and $k_{2,p}$ are the retention ratio and number for each view. 

Finally, the top $k_{2,p}$ tokens with the highest attention score $S_{p,i}^r$ to instruction tokens are selected to further enhance the task-alignment capability.

\section{Experiments}\label{Experiments}
This section first outlines the experimental setup, followed by a comparative evaluation against state-of-the-art (SOTA) methods. We then conduct ablation studies to validate the effectiveness of the proposed method. Finally, visualization results are presented to demonstrate its adaptability.

\renewcommand{\multirowsetup}{\centering}
\begin{table}[!ht]
    \centering
    \setlength{\tabcolsep}{4.0pt}
    \renewcommand{\arraystretch}{1.0}  
    \footnotesize
    \centering
    \caption{Comparison with SOTA methods on DriveLMM-o1 using DriveLMM-o1 model.}
    \scalebox{0.7}{
    \begin{tabular}{l c c c c c >{\centering\arraybackslash}p{2.1cm}}
        \toprule[1.5pt]
        \textbf{\begin{tabular}[c]{@{}c@{}}Methods \end{tabular}} &
  \textbf{\begin{tabular}[c]{@{}c@{}}Risk Assessment \\ Accuracy\end{tabular}} &
  \textbf{\begin{tabular}[c]{@{}c@{}}Traffic Rule\\  Adherence\end{tabular}} &
  \textbf{\begin{tabular}[c]{@{}c@{}}Scene Awareness \\ and Object Und.\end{tabular}} &
  \textbf{Relevance} &
  \textbf{\begin{tabular}[c]{@{}c@{}}Missing \\ Details\end{tabular}} &
  \textbf{\begin{tabular}[c]{@{}c@{}}Overall\\ Reasoning\end{tabular}} \\ \hline
          \rowcolor{gray!20} \multicolumn{7}{c}{\textit{Upper Bound, 256 Tokens per image} \ ${(100\%)}$}\\
          \hline

        {DriveLMM-o1} & {73.81} & {82.26} & {75.64} & {79.51} & {71.38} & \multirow{1}*{{74.37}} \\
        \hline

        \rowcolor{gray!20} \multicolumn{7}{c}{\textit{Retain 64 Tokens per image} \ ${(\downarrow 75\%)}$}\\
        \hline
        FastV\texttt{\scriptsize{(ECCV24)}} & 71.35 & 80.58 & 72.85 & 76.74 & 69.25 & 72.41\scriptsize{(97.4\%)} \\
        DART\texttt{\scriptsize{(EMNLP25)}} & 71.50& 78.21& 72.80& 77.33& 69.63 & 71.46\scriptsize{(96.1\%)} \\ 

        \multirow{1}*{SparseVLM\texttt{\scriptsize{(ICML25)}}} & 70.08 & 79.59 & 71.87 & 75.66 & 67.43 & 71.23\scriptsize{(95.8\%)} \\
        
        PACT\texttt{\scriptsize{(CVPR25)}} & 71.20 & 78.11 & 71.44 & 77.21 & 68.95 & 70.95\scriptsize{(95.4\%)}  \\
        DivPrune\texttt{\scriptsize{(CVPR25)}} & 71.66 & 78.39 & 73.04 & 77.45 & 69.92 & 71.67\scriptsize{(96.4\%)}  \\
        Prune2Drive\texttt{\scriptsize{(CVPR26)}} & {71.31} & {80.74} & {73.07} & {77.48} & {69.30} & {72.69}\scriptsize{(97.7\%)} \\

        \rowcolor{orange!20} \textbf{Ours}  & \textbf{72.91} & \textbf{81.96} & \textbf{74.85} & \textbf{78.55} & \textbf{71.58} & \textbf{73.79\scriptsize{(99.2\%)}} \\
        
        \hline
        \rowcolor{gray!20} \multicolumn{7}{c}{\textit{Retain 25 Tokens per image} \ ${(\downarrow 90\%)}$}\\
        \hline
        FastV\texttt{\scriptsize{(ECCV24)}} & 67.74 & 78.07 & 69.37 & 73.24 & 66.02 & 69.56\scriptsize{(93.5\%)} \\
        DART\texttt{\scriptsize{(EMNLP25)}} & 68.60& 75.81& 69.97& 74.94&67.60& 69.01\scriptsize{(92.8\%)} \\ 
        \multirow{1}*{SparseVLM\texttt{\scriptsize{(ICML25)}}} & 68.18 & 77.81 &69.87& 73.98& 66.20& 69.40\scriptsize{(93.3\%)} \\ 

        PACT\texttt{\scriptsize{(CVPR25)}} & 68.20& 76.23& 69.11& 74.14&66.10& 68.38\scriptsize{(91.9\%)} \\
        DivPrune\texttt{\scriptsize{(CVPR25)}} & 69.32& 76.47& 70.78& 75.53&68.22& 69.65\scriptsize{(93.7\%)} \\
        Prune2Drive\texttt{\scriptsize{(CVPR26)}} & {{70.76}} & {77.66} & {72.29} & {77.01} & {68.94} & {70.31}\scriptsize{(94.5\%)} \\
        
        \rowcolor{orange!20} \textbf{Ours}  & {\textbf{72.05}} & \textbf{81.81} & \textbf{73.80} & \textbf{77.61} & \textbf{70.25} & \textbf{72.91\scriptsize{(98.0\%)}} \\
        \bottomrule[1.5pt]
	\end{tabular}}
    \label{tab:drivelmm}
\end{table}

\subsection{Experimental Setup}
\textbf{Models.} We verify the effectiveness of the proposed method by experiment on two specialist autonomous driving visual-language models: DriveMM \cite{huang2024drivemm} and DriveLMM-o1 \cite{ishaq2025drivelmm}. Specifically, DriveMM is built based on LLaVA-OneVision-7B \cite{li2024llava} and fine-tuned across various autonomous driving datasets, can process diverse data inputs and driving tasks. DriveLMM-o1 is built based on InternVL2.5-8B \cite{chen2024internvl} and fine-tuned in the DriveLMM-o1 dataset to enhance step-by-step reasoning ability in driving scenarios.

\textbf{Benchmarks.} We evaluate our method on three large-scale image-based autonomous driving VQA benchmarks: DriveLM \cite{sima2024drivelm}, DriveLMM-o1 \cite{ishaq2025drivelmm}, and MAPLM \cite{cao2024maplm}, as well as the video-based benchmark STSnu \cite{fruhwirth2025stsbench}. DriveLM is annotated on the NuScenes \cite{caesar2020nuscenes} dataset, including QA pairs of perception, prediction, and planning tasks with six-view images. DriveLMM-o1 is also annotated on NuScenes to advance step-by-step visual reasoning for autonomous driving with six-view images as input. MAPLM is tailored for advanced map and traffic scene understanding, supporting tasks such as scene classification, lane counting, and point cloud quality analysis based on three-view images and synchronized LiDAR point cloud maps. STSnu is proposed to evaluate the spatio-temporal reasoning capabilities of VLMs based on six-view videos. We follow the common metrics in each benchmark for fair comparison. Details are in the supplementary materials.

\textbf{Evaluation Metrics.} We follow the common metrics in each benchmark for fair comparison: (1) DriveLM \cite{sima2024drivelm} implements four evaluation metrics: accuracy, GPT score, language-based evaluation (BLEU-4, Rouge, and CIDEr), and match score. (2) DriveLMM-o1 \cite{ishaq2025drivelmm} adopts five driving-specific reasoning evaluation metrics: Risk Assessment Accuracy, Traffic Rule Adherence, Scene Awareness and Object Understanding, Relevance, and Missing Details, as well as Overall Reasoning. (3) MAPLM \cite{cao2024maplm} uses FRM (Frame-overall-accuracy) and QNS (Question-overall-accuracy). 
(4) STSnu \cite{fruhwirth2025stsbench} categorizes questions into four interaction types and evaluates them separately: (i) Ego: Which of the following options best describes the ego vehicle’s driving maneuver? (ii) Agent: Which of the following options best describes the driving behavior of the <reference to agent>? (iii) Ego-to-agent: Which of the following options best describes the ego vehicle’s driving behavior with respect to the <reference to agent>? (iv) Agent-to-agent: Which of the following options best describes <reference to agent 1>’s maneuver with respect to <reference to agent 2>?

\textbf{Comparison Methods.} We compare our method with several recent representative token pruning methods, including FastV \cite{chen2024image}, DART \cite{wen2025stop}, SparseVLM \cite{zhang2024sparsevlm}, PACT \cite{dhouib2025pact}, and DivPrune \cite{alvar2025divprune}; as well as Prune2Drive \cite{xiong2025prune2drive} which is specifically designed for multi-view VLMs. To ensure a fair comparison, we optimized the pruning configuration for Prune2Drive on each benchmark using its official open-source implementation. To assess pruning robustness, we evaluate all methods under two widely adopted pruning settings, where on average 75\% and 90\% of visual tokens are pruned. The pruning ratio is computed as the average token reduction across all layers of the LLM, following prior works \cite{chen2024image,wen2025stop,zhang2024sparsevlm,dhouib2025pact,alvar2025divprune,xiong2025prune2drive}. Additional experiments under more pruning settings are provided in the supplementary material.

\textbf{Implementation Details.} Our two-stage pruning is applied before layers 0 and 16, with average pruning ratios of $r$ and $r^2$, causing an exponential reduction.

\renewcommand{\multirowsetup}{\centering}
\begin{table}[!t]
    \centering
    \setlength{\tabcolsep}{5.0pt}
    \renewcommand{\arraystretch}{1.0}
    \footnotesize
    \centering
    \caption{Comparison with SOTA methods on DriveLM using DriveMM model.}
    \scalebox{0.77}{
    \begin{tabular}{l c c c c c c c}
        \toprule[1.5pt]
        \textbf{Model} & \textbf{Accuracy} & \textbf{Chatgpt} & \textbf{BLEU-4} & \textbf{Rouge} & \textbf{CIDEr} & \textbf{Match}  & {\textbf{Average}}\\
        \hline
         \rowcolor{gray!20} \multicolumn{8}{c}{\textit{Upper Bound, 729 Tokens per image} \ ${(100\%)}$}\\
         \hline
        {DriveMM} & {0.81} & {65.44} & {0.61} & {0.74} & {0.19} & {33.9} & {{59.1}} \\
        \hline

         \rowcolor{gray!20} \multicolumn{8}{c}{\textit{Retain 180 Tokens per image} \ ${(\downarrow 75\%)}$}\\
        \hline
        
        FastV& 0.77 & 63.95 & 0.54 & 0.72 & 0.15 & 32.8 & {56.6\scriptsize{(95.8\%)}} \\
        
       DART  & 0.78 & 64.02 & 0.56 & 0.73 & 0.18 & 33.6 & 57.2\scriptsize{(96.8\%)} \\
        {SparseVLM } & 0.79 & 63.74 & 0.57 & 0.73 & 0.19 & 33.4 & 57.4\scriptsize{(97.1\%)} \\
        PACT  & 0.78 & 64.24 & 0.56 & 0.73 & 0.19 & 33.3 & 57.3\scriptsize{(97.0\%)}  \\
        DivPrune  & 0.79 & 64.67 & 0.59 & 0.74 & 0.21 & 33.7 & 57.9\scriptsize{(98.0\%)}  \\
        {Prune2Drive } & \textbf{0.80} & {64.92} & {0.60} & \textbf{0.75} & {0.20} & {34.0} & {58.3\scriptsize{(98.6\%)}} \\
        \rowcolor{orange!20} \textbf{Ours} & {0.79} & \textbf{65.52} & \textbf{0.61} & \textbf{0.75} & \textbf{0.22} & \textbf{{34.3}} & \textbf{58.7\scriptsize{(99.3\%)}} \\
        
        \hline

         \rowcolor{gray!20} \multicolumn{8}{c}{\textit{Retain 72 Tokens per image} \ ${(\downarrow 90\%)}$}\\
        \hline
        
        FastV & 0.68 & 63.52 & 0.49 & 0.71 & 0.08 & 32.3 & {54.1\scriptsize{(91.5\%)}} \\
        
        DART & 0.69 & 64.20 & 0.51 & 0.71 & 0.12 & 32.0 & 54.7\scriptsize{(92.6\%)} \\
        {SparseVLM } & 0.75 & 63.38 & 0.52 & 0.72 & 0.15 & 32.9 & 55.9\scriptsize{(94.6\%)} \\
        PACT  & 0.76 & {64.76} & 0.54 & 0.73 & 0.15 & 32.9 & 56.8\scriptsize{(96.1\%)}  \\
        DivPrune  & 0.78 & 64.16 & 0.55 & 0.73 & \textbf{0.17} & 32.7 & 57.0\scriptsize{(96.4\%)}  \\
        {Prune2Drive} & {0.78} & 64.52 & {0.56} & \textbf{0.74} & {0.16} & {33.4} & {57.4\scriptsize{(97.1\%)}} \\
        \rowcolor{orange!20} \textbf{Ours} & \textbf{0.79} & \textbf{65.41} & \textbf{0.58} & \textbf{0.74} & \textbf{0.17} & \textbf{34.0} & \textbf{58.2\scriptsize{(98.5\%)}} \\

        \bottomrule[1.5pt]
    \end{tabular}}
    \label{tab:drivemm}
\end{table}

\subsection{Main Results}
This section presents comparative results between our method and SOTA methods on DriveLMM-o1, DriveLM, MAPLM and STSnu benchmarks.

\textbf{DriveLMM-o1.} The results are presented in \cref{tab:drivelmm}. Our method retains 98.0\% and 99.2\% of the original performance using only 10\% and 25\% of the visual tokens, respectively, outperforming SOTA methods. Notably, under the 90\% pruning setting, our method outperforms the second-best method Prune2Drive by a substantial margin of 2.6 on the Overall Reasoning metric, without any offline hyperparameter search.

\textbf{DriveLM.} The results in \cref{tab:drivemm} show that our method retains near-original performance with only 10\% and 25\% tokens, achieving 98.5\% and 99.3\% of the original model, respectively. Notably, it even surpasses the original model on Chatgpt, Rouge, CIDEr and Match under the 75\% pruning setting. Compared to Prune2Drive, our method yields consistent improvements of 0.8 and 0.4.

\textbf{MAPLM.} Results in \cref{tab:maplm} show that under the more challenging multi-sensor input setting, our method exhibits a substantially smaller accuracy drop than SOTA methods. With 90\% pruning, FastV, SparseVLM, and Prune2Drive degrade by 17.63, 15.84, and 7.43, respectively, whereas our method drops by only 3.94, remaining close to the original model. This indicates that our method can effectively identify critical information across diverse inputs and exhibits stability and strong generalization capability across tasks.

\begin{table}[!t]
    \centering
    \renewcommand{\arraystretch}{1.0}
    \footnotesize
    \centering
    \caption{Comparison with SOTA methods on MAPLM using DriveMM model.}
    \scalebox{0.77}{
    \begin{tabular} {>{\raggedright\arraybackslash}p{2.0cm} >{\centering\arraybackslash}p{1.8cm} >{\centering\arraybackslash}p{1.8cm}  >{\centering\arraybackslash}p{2.2cm} >{\centering\arraybackslash}p{1.8cm} >{\centering\arraybackslash}p{1.8cm}  >{\centering\arraybackslash}p{2.2cm}}

        \toprule[1.5pt]
        \textbf{Model} & \textbf{FRM} & \textbf{QNS}  & {\textbf{Average}}& \textbf{FRM} & \textbf{QNS}  & {\textbf{Average}}\\
        \hline
         \rowcolor{gray!20} \multicolumn{7}{c}{\textit{Upper Bound, 729 Tokens per image} \ ${(100\%)}$} \\
         \hline
        {DriveMM} & {58.60} & {85.87} & {{72.24}} & {58.60} & {85.87} & {{72.24}}\\
        \hline

        \rowcolor{gray!20} & \multicolumn{3}{c}{\textit{Retain 180 Tokens per image} \ ${(\downarrow 75\%)}$} & \multicolumn{3}{c}{\textit{Retain 72 Tokens per image} \ ${(\downarrow 90\%)}$}\\
        \hline

        FastV & 47.67  & 81.70 & {64.69\scriptsize{(89.5\%)}} & 33.07  & 76.15 & {54.61\scriptsize{(75.6\%)}} \\
        
       {DART } & 51.27 & 83.27 & 67.27\scriptsize{(93.1\%)} & 40.27 & 79.23 & 59.75\scriptsize{(82.7\%)}\\
        {SparseVLM} & 49.80 & 82.52 & 66.16\scriptsize{(91.6\%)}& 35.67 & 77.13 & 56.40\scriptsize{(78.1\%)}\\
        {PACT } & 50.87 & 82.50 & 66.69\scriptsize{(92.3\%)} & 30.87 & 74.78 & 52.83\scriptsize{(73.1\%)}\\
        {DivPrune } & 53.46 & 83.10 & 68.28\scriptsize{(94.5\%)} & 48.80 & 80.75 & 64.78\scriptsize{(89.7\%)}\\

        {Prune2Drive } & {53.47} & {83.10} & {68.29\scriptsize{(94.5\%)}} & {48.33} & {81.28} & {64.81\scriptsize{(89.7\%)}}\\

        \rowcolor{orange!20} \textbf{Ours} & \textbf{56.07} & \textbf{84.55} & \textbf{70.31\scriptsize{(97.3\%)}}  &  \textbf{53.27} & \textbf{83.33} & \textbf{68.30\scriptsize{(94.5\%)}}\\

        \bottomrule[1.5pt]
    \end{tabular}}
    
    \label{tab:maplm}
\end{table}

\renewcommand{\multirowsetup}{\centering}
\begin{table*}[!t]
    \centering
    \caption{Comparison with SOTA methods on STSnu using DriveMM model.}

    \setlength{\tabcolsep}{4.5pt}
    \renewcommand{\arraystretch}{1.0}  
    \footnotesize
    \centering
    \scalebox{0.77}{
    \begin{tabular}{l c c c c c }
        \toprule[1.5pt]
        \textbf{\begin{tabular}[c]{@{}c@{}}Methods \end{tabular}} &
  \textbf{Ego} &
  \textbf{Ego-to-Agent} &
  \textbf{Agent} &
  \textbf{Agent-to-Agent} &
  \textbf{Average} \\ \hline
          \rowcolor{gray!20} \multicolumn{6}{c}{\textit{Upper Bound, 196 Tokens per image} \ ${(100\%)}$}\\
          \hline

        {DriveMM} & {43.16} & {59.33} & {43.70} & {30.67} & {44.21}\\
        \hline

        \rowcolor{gray!20} \multicolumn{6}{c}{\textit{Retain 49 Tokens per image} \ ${(\downarrow 75\%)}$}\\
        \hline
        FastV  & 40.02 & \textbf{63.08} & 42.72 & 26.81 & 43.16\scriptsize{(97.6\%)} \\
        DART  & 38.70& 58.32& \textbf{44.24}& 28.79& 42.51\scriptsize{(93.9\%)} \\ 
        \multirow{1}*{SparseVLM } & 31.04 & 59.62 & 43.29 & \textbf{32.89} & 41.71\scriptsize{(94.3\%)} \\
        
        PACT & 31.28 & {59.98} & 42.79 & 31.63 & 41.42\scriptsize{(93.7\%)}  \\
        DivPrune& 36.65 & 58.16 & 42.60 & 29.82 & 41.81\scriptsize{(94.6\%)} \\
        Prune2Drive & {36.65} & {{58.52}} & {43.04} & {30.51} & {42.18\scriptsize{(95.4\%)}} \\
        \rowcolor{orange!20} \textbf{Ours}  & \textbf{44.59} & {58.34} & {43.08} & {30.99} & \textbf{44.25\scriptsize{(100\%)}}\\
        
        \hline
        \rowcolor{gray!20} \multicolumn{6}{c}{\textit{Retain 19 Tokens per image} \ ${(\downarrow 90\%)}$}\\
        \hline
        FastV & 19.55 & 61.90 & 38.33 & 28.19 & 36.99\scriptsize{(83.7\%)} \\
        DART& 15.41& 37.76& 36.08& 26.35&28.90\scriptsize{(65.4\%)} \\ 
        \multirow{1}*{SparseVLM } & 26.06 & 69.18 &40.30& 29.16& 41.18\scriptsize{(93.1\%)}\\ 

        PACT& 15.86& 65.96& 39.00& 27.79&37.15\scriptsize{(84.0\%)}\\
        DivPrune & 30.75& 59.50& \textbf{43.79}& 30.95&41.25\scriptsize{(93.3\%)} \\
        Prune2Drive & {31.89} & {\textbf{72.76}} & {40.20} & {27.84} & {\textbf{43.17\scriptsize{(97.6\%)}}}\\
        
        \rowcolor{orange!20} \textbf{Ours}  & \textbf{39.33} & {57.75} & {43.49} & \textbf{31.46} & {43.00\scriptsize{(97.3\%)}} \\

        \bottomrule[1.5pt]
	\end{tabular}}
	
    \label{tab:STSnu}
\end{table*}

\textbf{STSnu.} On the STSnu benchmark, we apply our MVPruner and other methods independently to each frame of the six-view video sequences. The results are presented in \cref{tab:STSnu}. Under the 75\% pruning setting, our method even surpasses the original model by 0.04, outperforming SOTA methods and demonstrating its effectiveness in removing spatiotemporal redundancy.

By preserving information aligned with the dynamic requirements of the model’s reasoning process, our method achieves consistent performance gains across diverse benchmarks and exhibits strong generalization.

\textbf{Inference Efficiency.} We analyze the efficiency of our method using speedup in the prefilling and decoding stages, FLOPs, GPU peak memory usage and throughput, under the 90\% pruning setting. The results are shown on \cref{tab:efficiency}, our method achieves the fastest decoding time, the lowest FLOPs, and GPU peak memory usage. Despite achieving efficiency comparable to Prune2Drive, our method preserves performance more effectively and can be directly applied to diverse models and scenarios without additional adjustments or costly hyperparameter tuning, offering a more practical and scalable solution. 

We further profile each component using DriveMM on DriveLM with six views and 729 tokens/view under the 90\% pruning setting. As shown in \cref{tab:efficiency_mvpruner}, CCTS dominates the overhead due to pairwise similarity computation, taking 26.48 ms/view. The total pruning overhead is 163.9 ms, only 2.9\% of the full model’s end-to-end latency (5580 ms). The detailed computing budget of our method is detailed in the supplementary materials.

\begin{table}[!t]
    \centering
    \setlength{\tabcolsep}{5.2pt}
    \renewcommand{\arraystretch}{1.0}
    \footnotesize
    \caption{Efficiency comparison on DriveLMM-o1, DriveLM, and MAPLM benchmarks under the 90\% pruning setting. }
    \scalebox{0.67}{
    \begin{tabular}
    {@{} >{\centering\arraybackslash}p{0.8cm} |lccccccc} 

        \toprule[1.2pt] {\multirow{2}{*}{{\textbf{}}}}& 
        \multirow{2}{*}{\textbf{Methods}} & 
        \multicolumn{2}{c}{\textbf{Speedup (ms) $\downarrow$}} & 
        \multirow{2}{*}{\textbf{FLOPs $\downarrow$}} & 
        {\textbf{GPU Peak $\downarrow$}} & {\textbf{Throughput $\downarrow$}} &
        \multirow{2}{*}{\textbf{Performance $\uparrow$}} \\
           & &  \textbf{(Prefilling)} & \textbf{(Decoding)} & & \textbf{Memory (GB)} & \textbf{(samples/s)} &\\ 
        \midrule

        {\multirow{4}*{\rotatebox{90}{{\scriptsize DriveLMM-o1}}}} &\cellcolor{gray!20} {DriveLMM-o1} &\cellcolor{gray!20} 591 &\cellcolor{gray!20} 51.9 &\cellcolor{gray!20} {100\%} &\cellcolor{gray!20} {17.31}&\cellcolor{gray!20} 11.84 &\cellcolor{gray!20} {74.37} \\
         & + FastV  & 128.5\scriptsize{($\uparrow$4.60$\times$)} & 45.9\scriptsize{($\uparrow$1.13$\times$)} & 20.3\% & 16.61 & 10.57\scriptsize{($\uparrow$1.12$\times$)} & 69.56 \\
         & + SparseVLM  & 164.6\scriptsize{($\uparrow$3.59$\times$)} & 47.6\scriptsize{($\uparrow$1.09$\times$)} & 20.7\% & 16.65 & 9.55\scriptsize{($\uparrow$1.24$\times$)} & 69.40 \\
         
         & + Prune2Drive  & \textbf{127.1\scriptsize{($\uparrow$4.65$\times$)}} & 47.2\scriptsize{($\uparrow$1.10$\times$)} & {20.3\%} & 16.58 & \textbf{8.58\scriptsize{($\uparrow$1.38$\times$)}} & 70.31 \\
         
         & \cellcolor{orange!20}+ \textbf{Ours} &\cellcolor{orange!20} 155.5\scriptsize{($\uparrow$3.80$\times$)} &\cellcolor{orange!20} \textbf{42.2\scriptsize{($\uparrow$1.23$\times$)}} &\cellcolor{orange!20} \textbf{20.0\%} & \cellcolor{orange!20}\textbf{16.08} & 8.60\cellcolor{orange!20}\scriptsize{\textbf{($\uparrow$1.38$\times$)}} & \cellcolor{orange!20}\textbf{72.91} \\
        \midrule
        \midrule
        
         {\multirow{5}*{\rotatebox{90}{\scriptsize DriveLM}}} &\cellcolor{gray!20} {DriveMM} &\cellcolor{gray!20} {1968} &\cellcolor{gray!20} 65.3 &\cellcolor{gray!20} {100\%} &\cellcolor{gray!20} {16.47}&\cellcolor{gray!20} 8.12 &\cellcolor{gray!20} {59.1} \\
         & + FastV & 340.5\scriptsize{($\uparrow$5.78$\times$)} & 63.4\scriptsize{($\uparrow$1.03$\times$)} & 13.3\% & 15.99& 5.56\scriptsize{($\uparrow$1.46$\times$)} & 55.4 \\
         & + SparseVLM & 484.7\scriptsize{($\uparrow$4.06$\times$)} & 64.0\scriptsize{($\uparrow$1.02$\times$)} & 14.4\% & 16.00 & 5.88\scriptsize{($\uparrow$1.38$\times$)} & 55.9 \\
         
         & + {Prune2Drive} & \textbf{307.5\scriptsize{($\uparrow$6.40$\times$)}} & 59.9\scriptsize{($\uparrow$1.09$\times$)} & {13.4\%} & 16.00 & \textbf{5.49\scriptsize{($\uparrow$1.48$\times$)}} & {57.4} \\
         
         &\cellcolor{orange!20}+ {\textbf{Ours}} & \cellcolor{orange!20}395.6\scriptsize{($\uparrow$4.97$\times$)} & \cellcolor{orange!20}\textbf{51.8\scriptsize{($\uparrow$1.26$\times$)}} &\cellcolor{orange!20} \textbf{12.7\%} &\cellcolor{orange!20} \textbf{15.93} & \cellcolor{orange!20}{5.58\scriptsize{($\uparrow$1.45$\times$)}} & \cellcolor{orange!20}\textbf{58.2} \\
        \midrule
        \midrule
        
        {\multirow{5}*{\rotatebox{90}{\scriptsize MAPLM}}} &\cellcolor{gray!20} {DriveMM} &\cellcolor{gray!20} 1182 &\cellcolor{gray!20} 61.7 &\cellcolor{gray!20} {100\%} &\cellcolor{gray!20} {16.24}&\cellcolor{gray!20} 1.72 &\cellcolor{gray!20} {72.24} \\
        
         & + FastV & 286.9\scriptsize{($\uparrow$4.12$\times$)} & 58.2\scriptsize{($\uparrow$1.06$\times$)} & 15.6\% & 15.91 & 0.82\scriptsize{($\uparrow$2.10$\times$)} & 54.61 \\
         & + SparseVLM & 353.9\scriptsize{($\uparrow$3.34$\times$)} & 59.9\scriptsize{($\uparrow$1.03$\times$)} & 15.8\% & 15.93 & 1.11\scriptsize{($\uparrow$1.55$\times$)} & 56.40 \\
         & + {Prune2Drive} & \textbf{274.9\scriptsize{($\uparrow$4.30$\times$)}} & 55.6\scriptsize{($\uparrow$1.11$\times$)} & {15.6\%} & 15.91 & 0.69\scriptsize{($\uparrow$2.49$\times$)} & {64.81} \\
         
          &\cellcolor{orange!20}+ {\textbf{Ours}} & \cellcolor{orange!20}299.1\scriptsize{($\uparrow$3.95$\times$)} & \cellcolor{orange!20}\textbf{51.6\scriptsize{($\uparrow$1.20$\times$)}} &\cellcolor{orange!20} \textbf{15.2\%} & \cellcolor{orange!20}\textbf{15.88} &\cellcolor{orange!20} \textbf{0.64\scriptsize{($\uparrow$2.69$\times$)}} & \cellcolor{orange!20}\textbf{68.30} \\
         
        \bottomrule[1.2pt]
    \end{tabular}}
    
    \label{tab:efficiency}
\end{table}
\begin{table}[!t]
    \centering
    \caption{Latency analysis of each pruning module on DriveLM benchmark under the 90\% pruning setting.}
    \setlength{\tabcolsep}{4.5pt}
    \footnotesize
    \centering
    \scalebox{1.0}{
    \begin{tabular}{ccccc}
        \toprule[1.0pt]
        \textbf{\small{DRA}} & \textbf{CCTS (/view)} & \textbf{IRA} & \textbf{ITS (/view)} & \textbf{Total}\\ 
        \hline
        3.55ms & 26.48ms & 1.11ms & 0.06ms & 163.9ms \\
        \bottomrule[1.0pt]
	\end{tabular}}
    \label{tab:efficiency_mvpruner}
\end{table}

\begin{figure}[tbp]
\centering
\includegraphics[scale=0.21]{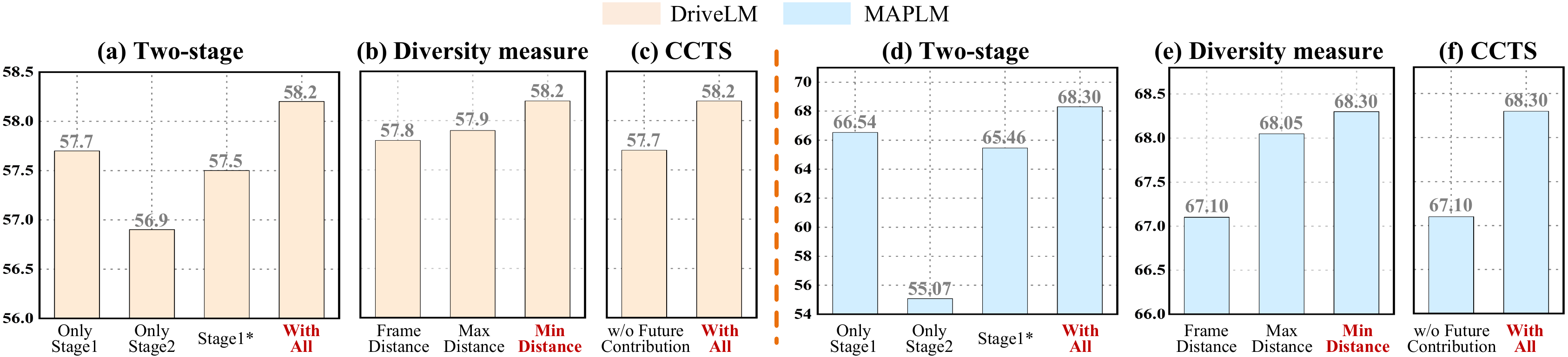}
\caption{Ablation studies on DriveLM and MAPLM benchmarks using DriveMM model.}
\label{fig:ablation}
\end{figure}

\subsection{Ablation study}
\label{main:ablation}
In this section, we conduct ablation studies on DriveLM and MAPLM under the 90\% pruning setting to evaluate the key design choices of our method.

\textbf{Two-stage Strategy.} We validate the proposed two-stage pruning strategy. Specifically, each stage is independently applied to the shallow layers to meet the target pruning ratio. Additionally, a variant in which stage 2 adopts the same strategy as stage 1 is evaluated, denoted as Stage1*, the results are shown in Figs. \ref{fig:ablation}(a, d). Applying stage 1 alone effectively removes redundant information while maximizing the semantic coverage of the retained token subset, satisfying the diversity-driven information requirements of shallow layers and outperforming SOTA methods even when used independently. In contrast, applying stage 2 independently yields inferior results, as early layers attend broadly to visual content and have not yet focused on task-relevant regions. Stage1* exhibits accuracy decline because the textual and visual modalities have already been aligned in deeper layers, and the model needs task-relevant information. These results confirm the necessity of the two-stage design, which aligns with the dynamic information requirements of the model during the reasoning process.

\begin{figure}[tbp]
\centering
\includegraphics[scale=0.26]{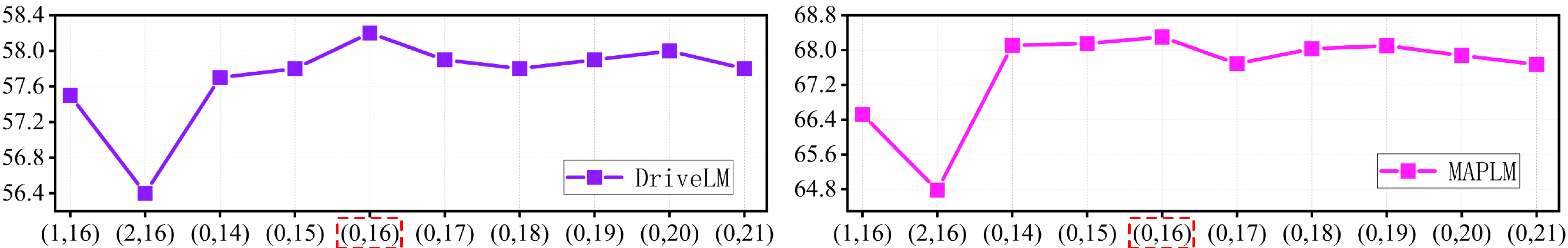}
\caption{Performance of different pruning layer selection strategies on DriveLM and MAPLM benchmarks using DriveMM model.}
\label{fig:ablation_layer}
\end{figure}

\begin{figure}[tbp]
\centering
\includegraphics[scale=0.22]{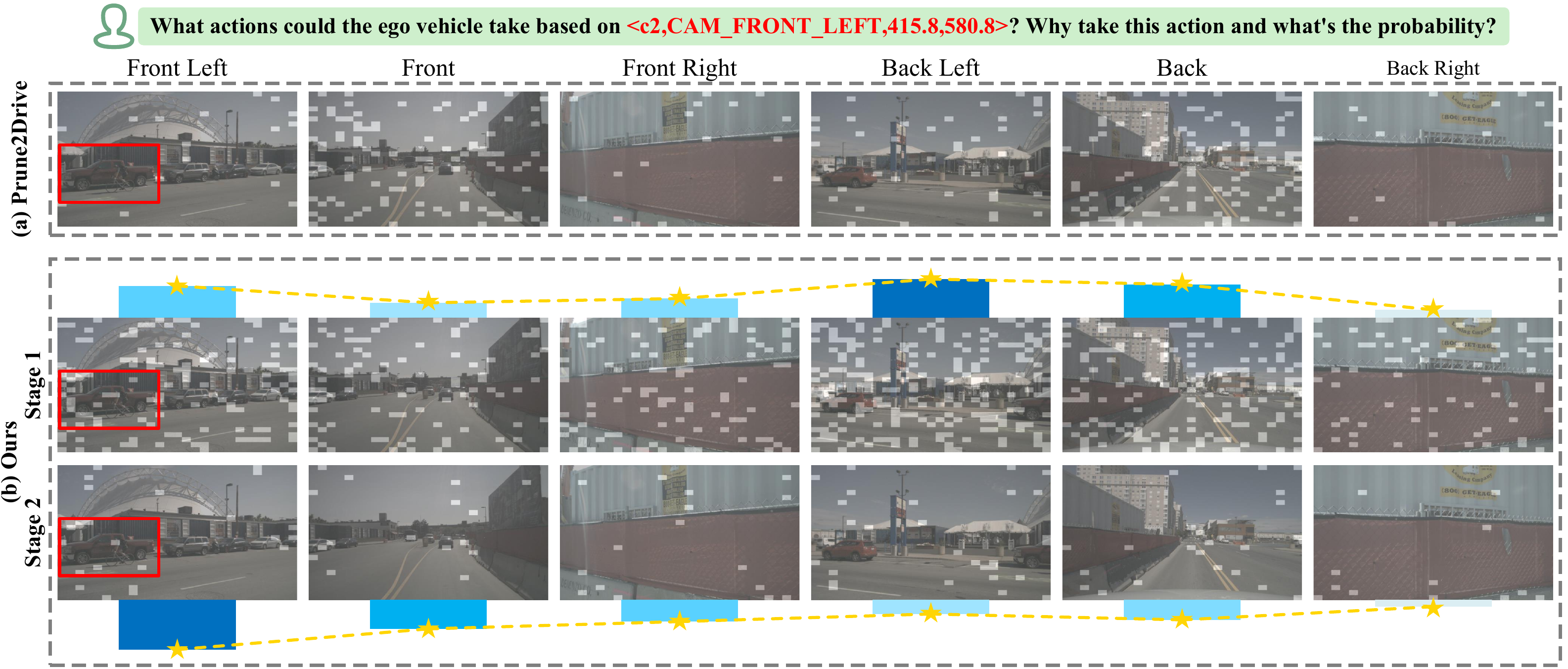}
\caption{\textbf{Quantitative comparison of kept tokens.} Taller and darker bars indicate view importance, where MVPruner allocates more tokens to information-rich views in stage 1 and to task-relevant views in stage 2.}
\label{fig:quantitative}
\end{figure}

\textbf{Diversity Measure in DRA.} We compare different diversity metrics for visual tokens in DRA, including distance to the global frame feature of the view, maximum pairwise distance, and minimum pairwise distance which we adopted. As shown in Figs. \ref{fig:ablation}(b, e), the choice of diversity metrics has only a minor impact on overall performance. However, the minimum pairwise distance consistently provides the most effective measure of token uniqueness.

\textbf{Metrics in CCTS.} We evaluate the impact of the future contribution metric on the CCTS. As shown in Figs. \ref{fig:ablation}(c, f), removing the future contribution metric causes the premature loss of potentially task-related information and degrades accuracy. By assessing each token’s contribution throughout the entire reasoning process, our method maximizes the representational capacity of the retained tokens in the shallow layer.

\textbf{Pruned Layers Selection.} We provide a detailed analysis of the pruning layer selection strategies, the results are shown in \cref{fig:ablation_layer}. The placement of stage 1 has an impact on the model, exhibiting a trend of performance degradation as the layer depth increases. One possible explanation is that, as the layer deepens, visual features progressively align with the text modality, resulting in the loss of fine-grained visual information and consequently reducing the reliability of the diversity metric. In contrast, the placement of stage 2 has only a minor effect on accuracy, indicating that the model in deeper layers can recognize task-related regions. Notably, the layer at which stage 2 achieves peak accuracy aligns with the layer exhibiting the highest task-related view recognition accuracy in Sec. \ref{main:analyze}, indicating that our prior analysis accurately captures the model’s ability to identify task-relevant information.

\begin{figure}[tbp]
\centering
\includegraphics[scale=0.22]{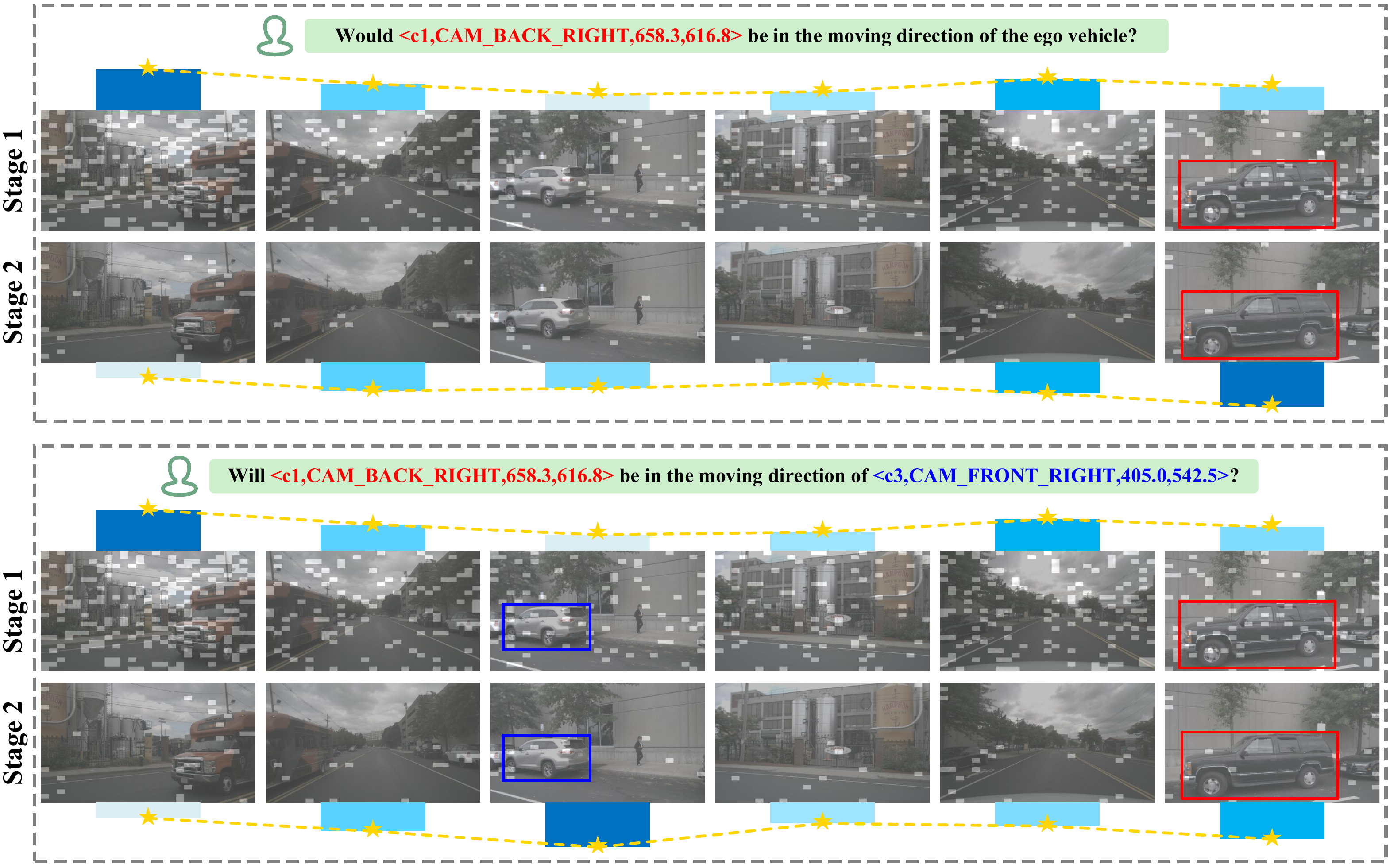}
\caption{\textbf{Visualization of kept tokens.} Taller and darker bars indicate view importance. Our method adaptively adjusts pruning budget allocation in response to variations in the instruction.}
\label{fig:quantitative_2}
\end{figure}

\subsection{Qualitative Analysis}
The qualitative comparison results between our method and Prune2Drive are shown in \cref{fig:quantitative}. Since most samples in the calibration set focus on front and back views, Prune2Drive allocates fixed, larger token budgets to them. Consequently, it fails to preserve relevant information when the model must attend to objects in the front-left view. In contrast, our method dynamically allocates the token budget across different reasoning stages according to the scene and task instructions. It maximizes the semantic representational capacity of retained tokens in shallow layers to provide comprehensive contextual information for subsequent layers, and precisely preserves task-relevant tokens in deeper layers.

\cref{fig:quantitative_2} shows samples with a fixed scene and varying instructions. Our method captures the resulting shifts in view importance and adaptively reallocates budgets, demonstrating strong generalization capability. More visualizations are in the supplementary materials.

\section{Conclusion}

In this paper, we analyze the ability of multi-view VLMs to identify important views and the underlying mechanisms. Results reveal that attention scores can effectively indicate task-related views in deeper layers and the reasoning process of VLMs exhibits dynamic information requirements. Guided by these findings, we propose MVPruner, which dynamically adjusts pruning ratios across views and retains the most critical tokens at each stage, aligning with the model’s dynamic information requirements. Extensive experiments demonstrate that our method achieves a superior efficiency–performance trade-off across diverse benchmarks and exhibits strong generalization.

\section*{Acknowledgments}
This work is supported in part by the National Natural Science Foundation of China under Grant U24B20127, in part by the National Key Research and Development Program of China under Grant 2024YFB4303102, and in part by the Chang'an University Graduate Student Research Innovation and Practice Program under Grant 300103267062.

%
%
\bibliographystyle{splncs04}
\bibliography{main}
\end{document}